\documentclass[10pt,twocolumn,letterpaper]{article}

\usepackage{cvpr}
\usepackage{times}
\usepackage{epsfig}
\usepackage{graphicx}
\usepackage{amsmath}
\usepackage{amssymb}
\usepackage{gensymb}
\usepackage{placeins}

\newcommand{\fig}[1]{Figure~\ref{fig:#1}}

\newcommand{\tab}[1]{Table~\ref{tab:#1}}

\newcommand{\eq}[1]{(\ref{eq:#1})}

\usepackage[pagebackref=true,breaklinks=true,letterpaper=true,colorlinks,bookmarks=false]{hyperref}

\cvprfinalcopy 

\def\cvprPaperID{5430} 

\ifcvprfinal\fi

\begin{document}

\title{Coordinate-based Texture Inpainting for Pose-Guided Human Image Generation}

\author{Artur Grigorev $^{1,2}$ \qquad Artem Sevastopolsky $^{1,2}$ \qquad Alexander Vakhitov $^1$ \qquad Victor Lempitsky $^{1,2}$\\
$^1$ Samsung AI Center, Moscow, Russia\\
$^2$ Skolkovo Institute of Science and Technology (Skoltech), Moscow, Russia\\
\texttt{\{a.grigorev,a.sevastopol,a.vakhitov,v.lempitsky\}@samsung.com}
}

\maketitle
\begin{abstract}
We present a new deep learning approach to pose-guided resynthesis of human photographs. At the heart of the new approach is the estimation of the complete body surface texture based on a single photograph. Since the input photograph always observes only a part of the surface, we suggest a new inpainting method that completes the texture of the human body. Rather than working directly with colors of texture elements, the inpainting network estimates an appropriate source location in the input image for each element of the body surface. This correspondence field between the input image and the texture is then further warped into the target image coordinate frame based on the desired pose, effectively establishing the correspondence between the source and the target view even when the pose change is drastic. The final convolutional network then uses the established correspondence and all other available information to synthesize the output image. A fully-convolutional architecture with deformable skip connections guided by the estimated correspondence field is used. We show state-of-the-art result for pose-guided image synthesis. Additionally, we demonstrate the performance of our system for garment transfer and pose-guided face resynthesis.

\end{abstract}

\section{Introduction}


\begin{figure*}
    \centering
    \includegraphics[width=0.9\textwidth]{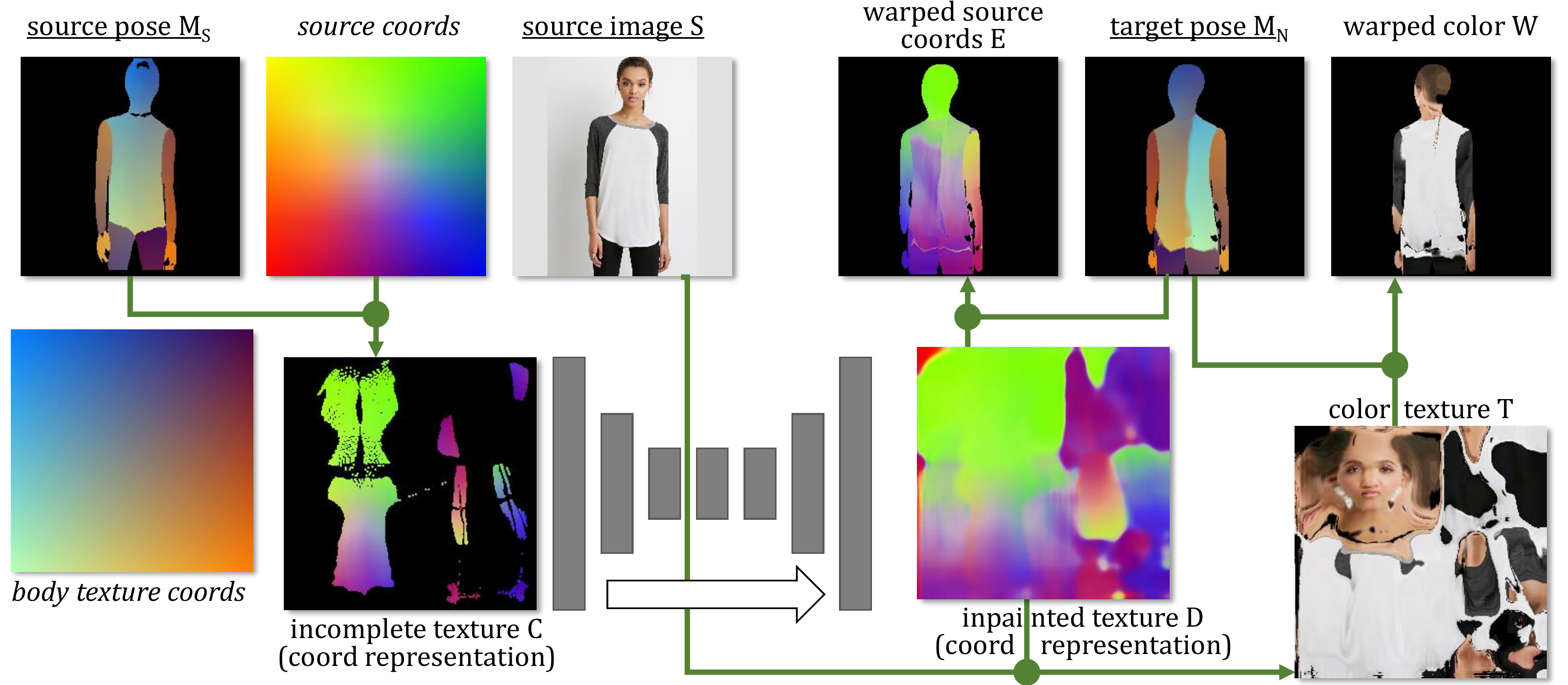}
    \caption{Coordinate-based texture inpainting. The scheme depicts the first (out of the two) part of our pipeline. Given the source pose (estimated by DensePose and converted to SMPL format), we rasterize the source coordinates of the known texture elements (e.g.\ by warping the source coordinate meshgrid). The resulting map is completed using deep convolutional network (gray) into a complete body texture, where for each texel a corresponding pixel coordinate in the source image is assigned. This correspondence map is then used to estimate the color texture. The second warping transforms the estimated texture maps into the target coordinate frame using the target pose, on which the resynthesis is conditioned (\textit{Data known at test time is underlined. 2D meshgrid arrays that define colormaps in the plot are in italic. Warping transforms are shown using green arrows, where the side connections correspond to the warp coordinates and straight arrows point from the data being warped}). }
    \label{fig:inpainting}
\end{figure*}

Learning human appearance from a single image (one-shot human modeling) has recently become an area of high research interest.  One interesting kind of the problem, which has a number of potential applications in augmented reality and retail, is pose-guided image generation~\cite{Ma17}. Here, the task is to resynthesize the view of a person from a new viewpoint and in a new pose, given a single input image. The progress in this problem benefits from the recent advances in human pose estimation and deep generative convolutional networks (ConvNets). A particular challenging setup considers humans wearing complex clothing, such as encountered in fashion photographs.

In this work we suggest a new approach for pose-guided person image generation. The approach is based on a pipeline that includes two deep generative ConvNets. The first convolutional network to estimate the texture of the human body surface from a small part of this texture (texture completion/inpainting). This texture is then warped to the new pose to serve as an input to the second convolutional network that generates the new view. 

One novelty of the approach lies in the texture estimation part (\fig{inpainting}), where the challenge is to utilize the natural symmetries of the human body. This task is non-trivial since the part of the texture that is known changes from one input image to another. As a result, straightforward image-to-image translation approaches result in very blurred textures, where the colors predicted at unknown locations are effectively averaged over very large number of input locations.

To solve this problem, we suggest a new method for texture completion, which we call \textit{coordinate-based texture inpainting}, and which results in a significant boost of the visual quality output for the entire pipeline. The method is based on a simple idea. Rather than working directly with colors of texture elements, the inpainting network works with coordinates of the texture elements in the source view. These values are analyzed by the inpainting network and then extended into the unknown part of the texture, so that each unknown texture element gets assigned a coordinate in the source view. Thus, a correspondence between source pixels and all points on the body surface is estimated. Using the estimated correspondence, the colors of each texture element can be transferred from the source view. The inpainting thus happens in the coordinate-space, while the extraction of colors from the source image, which generates the final texture, happens \textit{after} the inpainting. As a result, the inpainted textures retain high-frequency details from the source images.

Given the detailed texture generated by the coordinate-based inpainting process, the next step of the pipeline warps both the color texture and the source image coordinate maps  according to the target pose (which similarly to \cite{Neverova18} is defined by the DensePose~\cite{Guler18} descriptor). The final stage of the pipeline takes the warped images along with the pose information and maps it to the target image using a deep fully-convolutional encoder-decoder architecture with skip connections. The input image is used in this translation network, while the warped source image coordinates obtained during the texture inpainting process, are used to route the deformable skip connections~\cite{Siarohin18}.

Our contribution is thus two-fold. First, we suggest the new texture completion method that allows to retain high-level texture details even under large uncertainty. Secondly, we present a pose-guided person image generation pipeline that utilizes this method in two ways (to inpaint texture and to guide deformable skip connections) in order to generate new views with high realism and abundant texture details. Our method is evaluated on the popular Deep Fashion dataset \cite{Liu16}, where it obtains good results outperforming prior art. Furthermore, we additionally demonstrate the efficacy of coordinate-based texture inpainting idea on the face texture inpainting task for in-the-wild new view synthesis of faces, using the 300-VW dataset~\cite{Shen15}. As a coda, we show that a small modification of our approach can successfully be used to perform garment transfer (virtual try-on) with convincing results.

\section{Related Work}


\paragraph{Warping-based resynthesis.} There is a strong interest in using deep convolutional networks for generating realistic images \cite{Goodfellow14,Dosovitskiy17}. In the resynthesis case, when new images are generated by the change of geometry and appearance of the input images, it has been shown that using warping modules greatly enhances the quality of the re-synthesized images \cite{Ganin16,Zhou16}. The warping modules in this case are based on the differentiable (backward) grid sampler layer, which was first introduced as a part of Spatial Transformer Networks (STN)~\cite{Jaderberg15}. A large number of follow-up works on resynthesis reviewed below have relied on backward sampler. Here we revisit this building block and advocate the use of forward warping module.

\paragraph{Neural human resynthesis.} Neural-based systems for transforming an input view of a person into a new view with modified pose has been suggested recently. The initial works \cite{Ma17,Ma18,Esser18} used encoder-decoder type of architectures in order to perform resynthesis. More recent works use warping models that redirect either raw pixels or intermediate activations of the source view~\cite{Tulyakov18,Siarohin18,Zanfir18,Neverova18}. Our approach falls into this category and is most related to \cite{Neverova18}, as it utilizes the DensePose parameterization \cite{Guler18} within the network, and to \cite{Tulyakov18} as we use the idea of deformable skip connections from \cite{Tulyakov18}. We compare our results to \cite{Neverova18,Tulyakov18} and additionally to \cite{Esser18} extensively.

\paragraph{Texture completion.} Image inpainting based on deep convolutional networks is attracting increasing attention at the moment. Special variants of convolutional architectures adapted to the presence of gaps in the input data include Sheppard Networks~\cite{Ren15}, Sparsity-Invariant CNNs~\cite{Uhrig17}, networks with Partial Convolutions~\cite{Liu18}, networks with Gated Convolutions~\cite{Yu18}. We use the latter variant for our texture inpainting network. Learning body texture inpainting has two specific parts that distinguish it from generic image inpainting. First, complete textures may not be easily available and it is desirable to devise a method that can be trained from partial images. Secondly, textures are spatially aligned and possess symmetry structures that can be exploited, which calls for special-purpose algorithms. We are aware of only a few works which address these challenges specifically. Thus, UV-GAN~\cite{Deng18} utilizes the main axial symmetry of a face by passing an image and its flipped copy to an inpainting ConvNet. The system in \cite{Zanfir18} estimates a matrix that corresponds to the probabilities of SMPL model vertices to have similar colors, and use it to color vertices with unobserved colors.


\paragraph{Garment transfer.} We also show that a small modification of our approach can be used to transfer clothing from the photograph of one person to the photograph of a different person in a different pose. Most existing works that utilize neural networks can only handle very limited amount of deformation between the source image and the target view \cite{Han17,Jetchev17,Wang18}. The only work that we are aware of that can handle similar amount of pose change is SwapNet~\cite{Raj18}, which however only present results at low resolution. We perform comparison to \cite{Raj18} in the experimental section.

\paragraph{Face resynthesis.} Our approach is related to a number of very recent face resynthesis works that operate by warping the input image into the output image. These works include deforming autoencoders~\cite{Shu18} and X2Face~\cite{Wiles18}. An older class of works going back to the seminal Blanz and Vetter morphable model~\cite{Blanz99} estimate face texture from its fragment using a parametric model.

\section{Methods}

\newcommand{\x}{\mathbf{x}}
\renewcommand{\u}{\mathbf{u}}
\renewcommand{\v}{\mathbf{v}}
\newcommand{\y}{\mathbf{y}}
\newcommand{\w}{\mathbf{w}}
\newcommand{\m}{\mathbf{m}}
\renewcommand{\a}{\mathbf{a}}
\newcommand{\bw}{\text{bw}}
\newcommand{\fw}{\text{fw}}
\newcommand{\gt}{\text{gt}}

\paragraph{Problem formulation.} Our goal is to synthesize the new view of the person $N$ from the source view $S$. The resynthesis progresses by estimating the texture $T$. Below, we use the indexing $[x,y]$ to denote locations in the image frame (both the source and the new view), and we use the indexing $[u,v]$ to denote locations in the texture. We refer to source and target image elements and locations as \textit{pixels}, and to texture elements and positions as \textit{texels}. 

The texture is linked with the source and the new views, and following \cite{Neverova18} we assume that both for the source and the new view a mapping from a subset of the pixels covering the body (excluding hair and loose clothing) to the body texture positions is known. We thus assume that for each pixel $[x,y]$ in the source image (respectively in the new image) exists a mapping $M_S[x,y]$ (respectively $M_N[x,y]$) that associates with $[x,y]$ a position $[u,v] = [M^1_S[x,y],M^2_S[x,y]]$ (respectively, $[u,v] = [M^1_N[x,y],M^2_N[x,y]]$) on the texture. For pixels $[x,y]$ that do not fall within the projection of the human body, the mappings $M_N$ and $M_S$ are undefined. 

We assume that $M_S[x,y]$ and $M_N[x,y]$ are given and our goal is thus to estimate the new unknown view $N$ given its body texture mapping $M_N[x,y]$, as well as the known source view $S$ and its body texture mapping $M_S$. 

\paragraph{Texture map format and output conditioning.} We use the SMPL texture format~\cite{Loper15}. To make our approach comparable with \cite{Neverova18}, we estimate the mappings $M_S$ and $M_N$ based on DensePose~\cite{Guler18}, and then convert them to SMPL coordinates using a predefined mapping (provided with the DensePose). Thus, unlike \cite{Neverova18}, we use a single body texture during transfer. The information that is used to encode the source and the target pose is however exactly the same (the DensePose encoding), making the methods directly comparable.

\begin{figure}
    \centering
    \setlength{\tabcolsep}{.1em}    
    \begin{tabular}{ccc}
    \includegraphics[trim={50 0 50 0},clip,height=3cm]{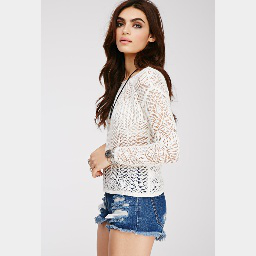}&
    \includegraphics[height=3cm]{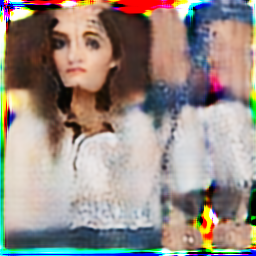}&
    \includegraphics[height=3cm]{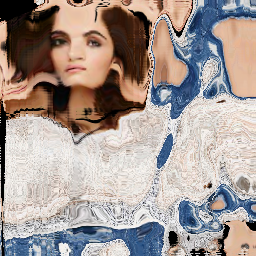}\\
    \includegraphics[trim={50 0 50 0},clip,height=3cm]{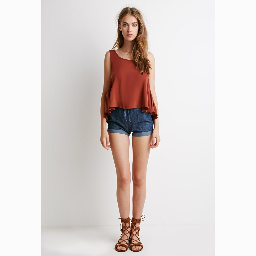}&
    \includegraphics[height=3cm]{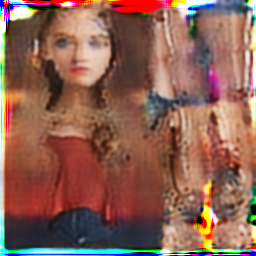}&
    \includegraphics[height=3cm]{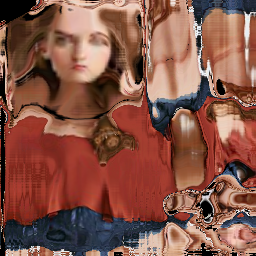}\\
    \includegraphics[trim={50 0 50 0},clip,height=3cm]{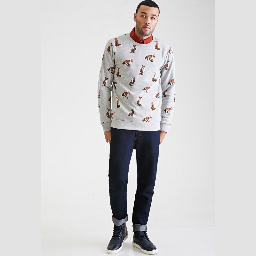}&
    \includegraphics[height=3cm]{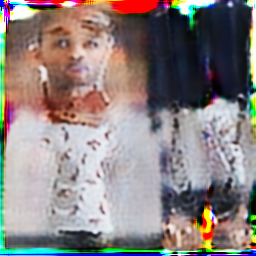}&
    \includegraphics[height=3cm]{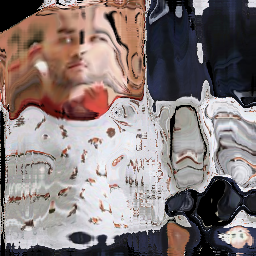}\\    
    \end{tabular}
    \caption{Body surface textures estimated using color-based inpainting (middle) and coordinate-based inpainting (right) for the inputs on the holdout set (left). Both inpaintings are generated using deep networks that were trained end-to-end with a variety of standard losses. Coordinate-based inpainting generates textures with more details leading to better final resynthesis results.}
    \label{fig:rgbvsxy}
\end{figure}

\paragraph{Coordinate-based texture inpainting.} The first step of our pipeline estimates the complete body surface texture from the source image $S$, and the mapping $M_S$. We first rasterize the source image coordinates over texture using warping. In more detail, we use scattered interpolation with bilinear kernel, so that each source pixel $[x,y]$ is rasterized at position $[M_S^1[x,y],M_S^2[x,y]]$. Unlike \cite{Neverova18}, we rasterize not the color values, but the values $x$ and $y$ themselves (in other words we apply scattered interpolation to the meshgrid array). The result of this warping step is the source coordinate map $C$, which for each texture element (texel) $[u,v]$ defines a corresponding location $[x,y] = [C^1[u,v],C^2[u,v]]$ in the source image. Since only a part of a human body can be visible in the source photograph, for a big part of texels, the source image location is undefined. When passing $C$ into the network, we set the unknown values to a negative constant (-10), and also provide the network with the mask $C'[u,v]$ of known texels.

The first learnable module of our pipeline is the inpainting network $f(C,C';\phi)$ with learnable parameters $\phi$ that takes an incomplete coordinate map $C$ in the texture space along with the mask of known texels, and outputs a completed and corrected source correspondence map $D$, where for each $[u,v]$ the corresponding location in the source image is defined:
\begin{equation} \label{eq:inpaint}
    D = f(C,C';\phi)\,.
\end{equation}
The mapping $f$ has a fully-convolutional structure. The task of the network is to learn the symmetries typical for human body and human dress, such as the left-right symmetry between body parts as well as less obvious symmetries. E.g.\ the network has a chance to learn that many clothings have repeated textures, so that if a guess needs to be made about the texture of the back from the front view, the best the network can do is to copy the frontal texture. Since the network $f$ deals with the inpainting task, we utilize the recently proposed gated convolution layers~\cite{Yu18} instead of standard convolutional layers. We use an hourglass (without skip-connection) architecture with 14 convolutional layers and 2.8 millions of parameters.

Given the estimated source correspondence map $D$, we can obtain the completed texture by sampling the original image using the locations prescribed by $D$:
\begin{equation} \label{eq:txsample}
    T[u,v] = S[D^1[u,v],D^2[u,v]]\,.
\end{equation}
where the bilinear sampling operator~\cite{Jaderberg15} is used to sample the source image at fractional locations. 


It is interesting to compare the way our approach (\textit{coordinate-based inpainting}) obtains the complete texture with the way the texture is obtained by other texture inpainting approaches (\textit{color-based inpainting}), e.g.\ \cite{Neverova18,Deng18,Zanfir18}. In the case of the color-based inpainting, the sampling \eq{txsample} and the inpainting operation \eq{inpaint} are swapped, i.e.\ the colors are first sampled from the source image to the texture leading to an incomplete color texture and then the incomplete color texture is inpainted using a learnable convolutional architecture. As we have compared the two approaches, we have found that due to a very high uncertainty and multimodality of the texture inpainting task, the color-based inpainting produces the textures with very blurred details as compared to the coordinate-based inpainting (see Fig.~\ref{fig:faces_ablation}). As will be shown in the experiments, when embedded into end-to-end resynthesis pipeline, considerably better results are obtained with coordinate-based inpaintings. 

\begin{figure}
    \centering
    \includegraphics[width=.5\textwidth]{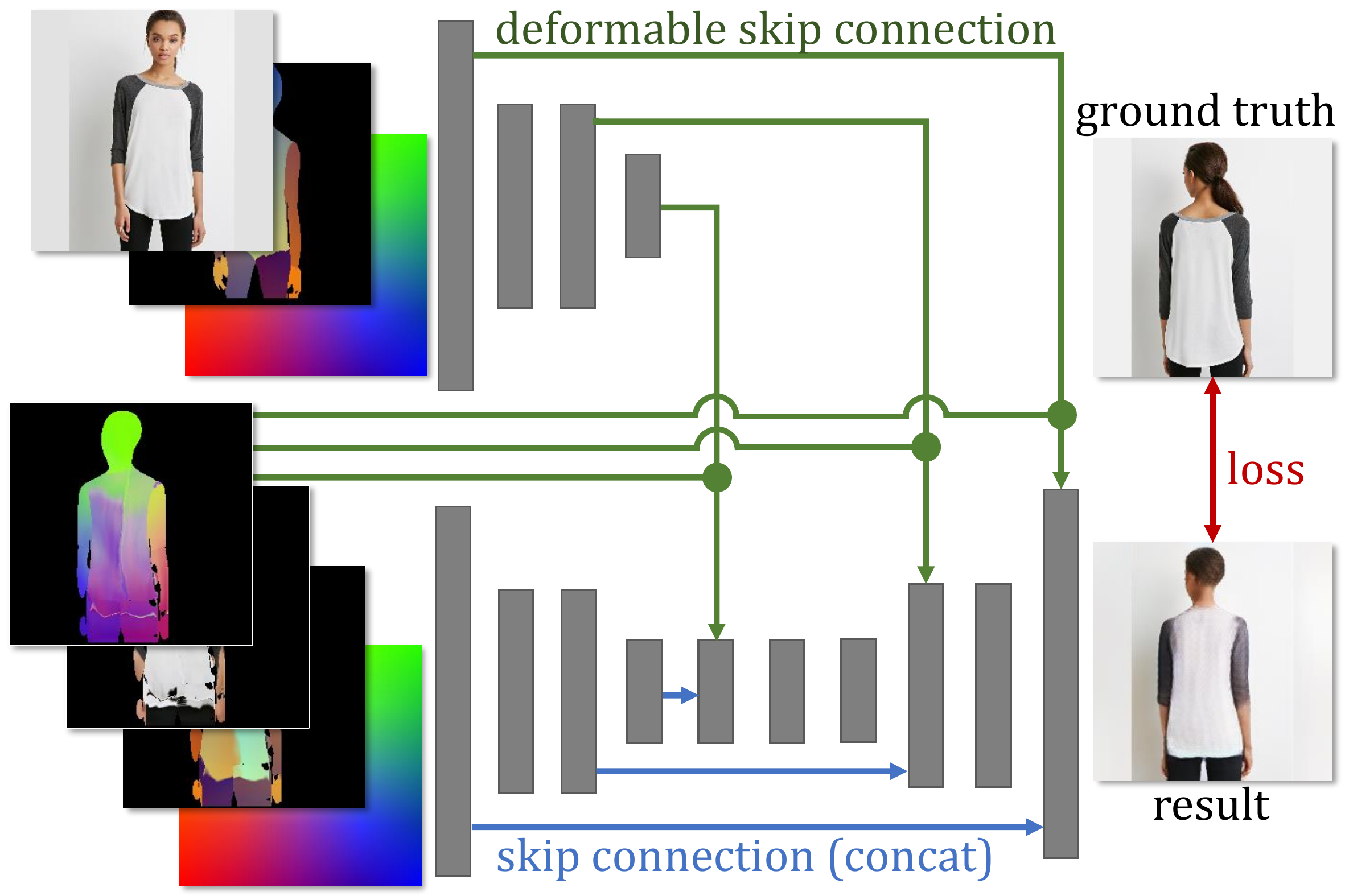}
    \caption{Final resynthesis. The second (of the two) part of our pipeline that takes the maps computed by the inpainting stage and map them to the final output image. Two separate encoders are used for maps aligned with the source pose (source pose, source image, meshgrid) and for maps aligned with the target pose (target pose, warped color texture, warped source coordinate map, meshgrid). The network has a U-Net type architecture (with intermediate residual blocks). Deformable skip connections are used to pass the activations of the source coordinate encoder to the joint decoder. The estimated correspondence map between the target and the source image is used to guide the deformable skip connections. Standard loss functions computed between the output of the pipeline and the ground truth target image in each pair are used for learning.}
    \vspace{-3mm}
    \label{fig:resynthesis}
\end{figure}

\paragraph{New view resynthesis.} Similarly to \cite{Neverova18}, in order to re-synthesize the target view, we warp the obtained color texture $T$ as well as the coordinate-based texture map $D$ to the new image frame, using the backward bilinear warping: 
\begin{align}
    W[x,y] = T\left[M_N^1[x,y],M_N^2[x,y]\right]\,,   \label{eq:colorwarp}\\ \pagebreak
    E[x,y] = D\left[M_N^1[x,y],M_N^2[x,y]\right]\,,   \label{eq:coordwarp}
\end{align}
where $W$ and $E$ are the new maps containing RGB color and the source view location for each body pixel of the target view. The values for non-body pixels are undefined (set to zeros in practice). The warping \eq{coordwarp} effectively estimates the correspondence between the target and the source views.

The final stage of our pipeline is a single convolutional network $g$ that converts (translates) the maps $W$, $E$, as well as the input maps $S$, $M_S$, and $M_N$ into an output image $N$. 
We first consider a straightforward architecture that takes all five maps, together with the meshgrid defined over the image frame as an input and uses the architecture of \cite{Johnson16} with added skip-connections to synthesize the output image. 

One caveat is that the input maps $S$, $M_S$ are not in any ways aligned with the target new image, which is known to cause problems.
As a more advanced variant (\fig{resynthesis}), we have used the deformable skip connections~\cite{Siarohin18} idea. Towards this end, we use a separate encoder part for the two maps $S$ and $M_S$ concatenated with a separate meshgrid. When passing the activations of this encoder into the decoder, we use the warp field $E$ and its downsampled versions to do bilinear resampling of the activations. In the experiments, we compare both variants of the architecture and find that deformable skip connections considerably boost the performance of our pipeline.

\begin{figure*}[ht!]

\setlength{\tabcolsep}{.1em}

\newcommand{\sbs}[2]{
            \includegraphics[trim={#2 0 #2 0},clip,height=2.0cm]{figures/sidebyside/#1_source.png} &
            \includegraphics[trim={#2 0 #2 0},clip,height=2.0cm]{figures/sidebyside/#1_target.png} &
            \includegraphics[trim={#2 0 #2 0},clip,height=2.0cm]{figures/sidebyside/#1_DSC.png} &
            \includegraphics[trim={#2 0 #2 0},clip,height=2.0cm]{figures/sidebyside/#1_DPT.png} &
            \includegraphics[trim={#2 0 #2 0},clip,height=2.0cm]{figures/sidebyside/#1_VUNET.png} &
            \includegraphics[trim={#2 0 #2 0},clip,height=2.0cm]{figures/sidebyside/#1_OURS_DP.png} &
            \includegraphics[trim={#2 0 #2 0},clip,height=2.0cm]{figures/sidebyside/#1_OURS_STICKMAN.png} 
        }

    \centering
    \begin{tabular}{cccccccccccccc}
         \sbs{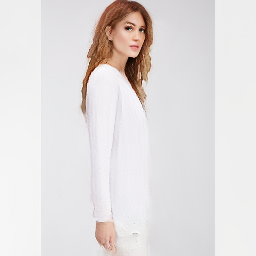}{50} & \sbs{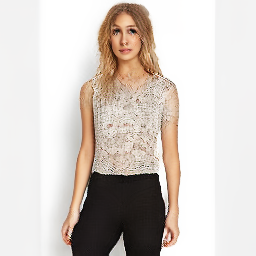}{50} \\
         \sbs{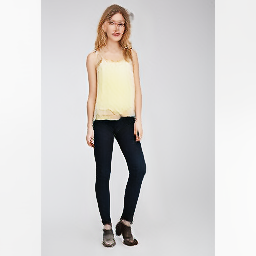}{50} & \sbs{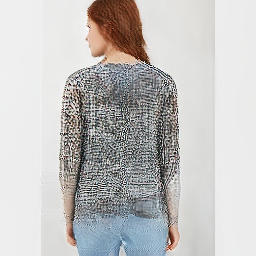}{50} \\
         \sbs{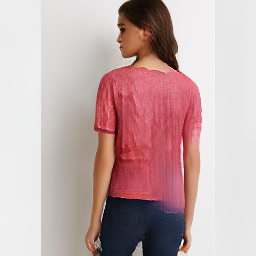}{50} & \sbs{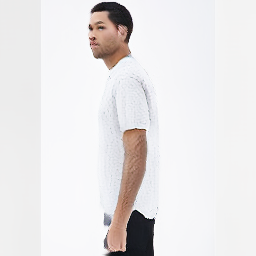}{50} \\
         \sbs{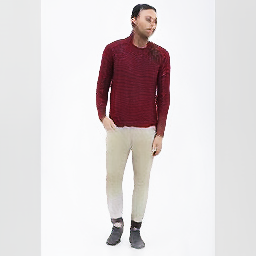}{50} & \sbs{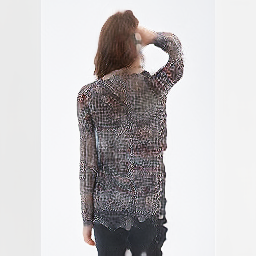}{50} \\
         SRC & GT & \cite{Siarohin18} & \cite{Neverova18} & \cite{Esser18} & Ours-D & Ours-K &
         SRC & GT & \cite{Siarohin18} & \cite{Neverova18} & \cite{Esser18} & Ours-D & Ours-K
    \end{tabular}
    \caption{Side-by-side comparison with state-of-the-art (first eight samples from the test set). We show source image (SRC), ground truth in the target pose (GT), deformable GAN~\cite{Siarohin18}, our method conditioned on dense pose~(Ours-D), and our method conditioned on keypoints (Ours-K). Consistently with the user study on a broader set, our method is more robust and has less artefacts than the state-of-the-art \cite{Siarohin18,Neverova18} on this subset. \textit{Electronic zoom-in recommended.} }
    \label{fig:sota}
\end{figure*}



\paragraph{Training procedure.} Our complete pipeline includes two convolutional networks, namely the inpainting network $f$ that performs coordinate-based texture completion, and the final network $g$. Both networks are trained on quadruplets $\{S,M_S,N,M_N\}$.
We first train the network $f$ by minimizing the loss comprising two terms: (1) the $\ell_1$ difference between the input incomplete texture $C$ and the inpainted texture $D$, where the difference is computed over texels that are observed in $C$; (2) the $\ell_1$ difference between the inpainted texture $D$ and the incomplete output texture that is obtained by warping the target image $N$ into the texture space using the map $M_N$, where the difference is computed over texels that are observed in the output image. 

After that, we fix $f$ and optimize the weights of network $g$, where we minimize the loss between the predicted $\hat{N}$ and the ground truth new view $N$. Here, we combine the perceptual loss~\cite{Johnson16} based on the VGG-19 network~\cite{Simonyan14}, the style loss~\cite{Gatys16} based on the same network, the adversarial loss~\cite{Goodfellow14} based on the patch GAN discriminator~\cite{Isola17} and the nearest neighbour loss introduced in~\cite{Siarohin18} (that proved to be a good substitution for $l_1$ loss used in \cite{Neverova18}). While the first network $f$ can be fine-tuned during the second stage, we did not find it beneficial for the resulting image quality.

\paragraph{Garment transfer.} A slight modification of our architecture allows it to perform garment transfer \cite{Han17,Jetchev17,Wang18,Raj18}. Here, given two views A and B, we want to synthesize a new view, where the pose and the person identity is taken from the view B, while the clothing is taken from view A. We achieve this by taking the architecture outlined above, and additionally conditioning the network $g$ on the masked image $N'$ of the target view, where we mask out all areas except face, hair, hat, and glasses.

The network $g$ is trained on the pairs of views of the same person, and effectively learns to copy heads and hands from $N'$ to $N$. 
At test time, we provide the network the identity-specific image $N'$ and the body texture mapping $M_N$ that are both obtained from the image of a different person from the one depicted in the input view. We show that our architecture successfully generalizes to this setting and thus accomplishes the virtual re-dress task.

\section{Applications and experiments}

\subsection{Pose-guided image generation}
\label{subsection:pose-guided_image_generation}

For the main experiments, we use the DeepFashion dataset (the in-shop clothes part)~\cite{Liu16}. In general, we follow the same splits as used in \cite{Siarohin18,Neverova18} that include 140,110 training and 8,670 test pairs, where clothing and models do not overlap between train and test sets.

\paragraph{Network architectures.} For the texture inpainting network $f$ we employ an hourglass architecture with gated convolutions from~\cite{Yu18} which proved effective in image reconstruction tasks with large hidden areas. The refinement network $g$ is also a hourglass network that has two encoders that map images by a series of gated convolutions interleaved with three downsampling layers resulting in $256\times 64 \times64$ feature tensors. This is followed by consecutive residual blocks and concluded by a decoder. The encoder and the decoder are also connected via three skip connections (at each of three resolutions). The encoder that works with $S$ and $M_S$ is connected to the decoder with deformable skip connections that are guided by the deformation field $E$. The network $f$ has 2,824,866 parameters, and the network $g$ has 11,382,984 parameters.


\paragraph{Comparison with state-of-the-art.} We compare the results of our method (full pipeline) with three state-of-the-art works~\cite{Neverova18,Siarohin18,Esser18}. We again follow the previous work~\cite{Neverova18} closely using structural self-similarity (SSIM) along with its' multi-scale version (MS-SSIM) metrics~\cite{Wang04} to measure the structure preservation and the inception score (IS)~\cite{Salimans16} to measure image realism. We also use recently introduced perceptual distance metric (LPIPS) \cite{Zhang2018} which measures distance between images using a network trained on human judgements (\tab{sota}).

Additionally  we perform a user study to compare our results with state-of-the-art based on $80$ image pairs from the test set (the indices of the pairs, as well as the results of \cite{Neverova18,Siarohin18,Esser18} were kindly provided by the authors of \cite{Neverova18}). In the user study, we have shown our results alongside of \cite{Neverova18,Siarohin18,Esser18} and asked to pick the variant, which was best fitting the ground truth (target) image. The source image was not shown. The order of presentation was normalized. 
50 people were involved in the user study. Each of them were to choose more realistic image in each of $80$ pairs. In \textbf{90\%} cases our reconstructions were preferred over those of~\cite{Neverova18} and in \textbf{76.7\%} cases cases over~\cite{Siarohin18}, while against~\cite{Esser18} our results were considered more realistic in \textbf{71.6\%} cases (approximately 4000 pairs were compared in each of the three cases).

\paragraph{Ablation study.} We evaluate the full variant of our approach that is described above, as well as the following ablations. In the \textit{Ours-NoDeform} ablation we do not use the deformable skip-connections in the network $f$, resulting in a single encoder for $W$, $E$, $S$, $M_S$, $M_N$ even though some of them ($S$, $M_S$) are aligned with the source view, while others ($W$, $E$, $M_N$) are aligned with the target view.

In the \textit{RGB inpainting} ablation we additionally replace coordinate-based inpainting with color-space inpainting, so that the output of the texture inpainting stage is only the color texture $T$, which is warped according to $M_N$ into the warped texture $W$ aligned with the target view. Since the map $E$ is unavailable in this scenario, no deformable skip-connections are used in this case. Finally, the \textit{No textures} ablation simply uses the maps $S$, $M_S$, and $M_N$ as an input to the translation network, ignoring texture estimation step altogether.

We compare the full version of the algorithm in terms of same four metrics: SSIM, MS-SSIM, IS and LPIPS. To ensure superiority of coordinate-based inpainting to color-based we have also performed a user study comparing \textit{Ours-Full} and \textit{RGB inpainting} methods. During this evaluation \textit{Ours-Full} were preferred in \textbf{62.7\%} cases.

\begin{figure*}[t]
    \setlength{\tabcolsep}{.1em}
    \centering
    \vspace{0.cm}
    \newcommand{\rdrs}[2]{
            \includegraphics[trim={#2 0 #2 0},clip,height=2.8cm]{figures/redress/#1_target.png} &
            \includegraphics[trim={#2 0 #2 0},clip,height=2.8cm]{figures/redress/#1_source.png} &
            \includegraphics[trim={#2 0 #2 0},clip,height=2.8cm]{figures/redress/#1_refined.png}  
        }

    \centering
    \begin{tabular}{ccccccccc}
         \rdrs{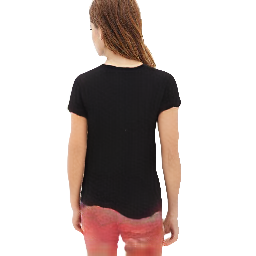}{50} \,\,& \rdrs{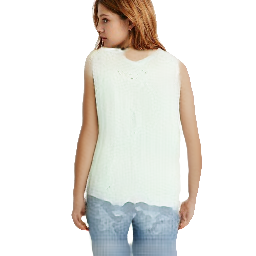}{50} \,\,&
         \rdrs{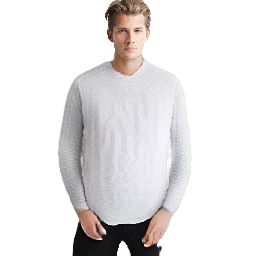}{50} \\ \rdrs{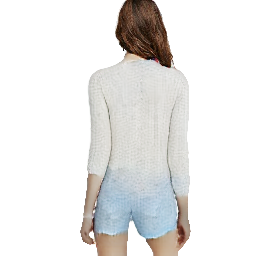}{50} \,\,&
         \rdrs{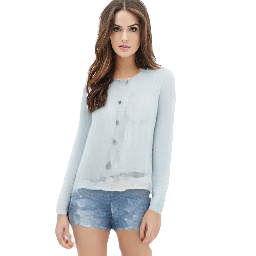}{50} \,\,& \rdrs{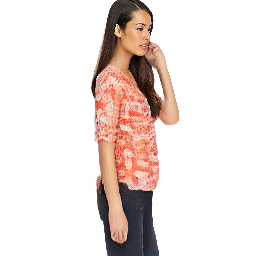}{50} \\
         Person & Cloth & Try-on &
         Person & Cloth & Try-on &
         Person & Cloth & Try-on 
    \end{tabular}
    \caption{Examples of garment transfer procedure obtained using a simple modification of our approach. In each triplet, the third image shows the person from the first image dressed into the clothes from the second image. }
    \label{fig:garment}
\end{figure*}


\begin{table}[]
    \centering
    \begin{tabular}{c|cccc}
                                     &  SSIM$\uparrow$ & MS-SSIM$\uparrow$ & IS$\uparrow$ & LPIPS$\downarrow$\\ \hline
         Ours & \textbf{0.791} & \textbf{0.810} & 4.46 & \textbf{0.169} \\ \hline
         DPT~\cite{Neverova18} &  0.785 & 0.807 & 3.61 & --- \\
         DSC~\cite{Siarohin18}&  0.761 & --- & 3.39 & --- \\
         VUnet~\cite{Esser18}&  0.753 & 0.757 & \textbf{4.55} & 0.196\\
    \end{tabular}
    \caption{Comparison with state-of-the-art. Our approach outperforms the other three in three of the four used metrics, although we found SSIM, MS-SSIM and IS to be much less adequate judgements of visual fidelity than user judgements. Arrows $\uparrow$, $\downarrow$ tell which value is better for the score — larger or smaller, respectively. Since we do not have access to full test set and code of some methods, values for metrics not presented in the respective papers are missing.}
    \label{tab:sota}
\end{table}

\paragraph{Keypoint-guided resynthesis.} It can be argued that our method (as well as \cite{Neverova18}) has an unfair advantage over \cite{Siarohin18,Esser18} and other keypoint-conditioned methods, since DensePose-based conditioning provides more information about the target pose compared to just keypoints (skeleton). To  address this argument, we have trained a fully-convolutional network that rasterizes the OpenPose~\cite{cao2017realtime}-detected skeleton over a set of maps (one bone per map) and train a network to predict the DensePose~\cite{Neverova18} result. We fine-tune our full network, while showing such ``fake'' DensePose results for the target image, effectively conditioning the system on the keypoints at test time. We add this variant to comparison  and observe that the performance of our network in this mode is very similar to the mode with DensePose conditioning (\fig{sota}). 

\paragraph{Garment transfer.} We also show some qualitative results of the garment transfer (virtual try-on). The garment transfer network was obtained by cloning our complete pipeline in the middle of the training and adding the masked target image (with revealed face and hair)  to the input of the network. During training background on ground truth targets is segmented out by the pretrained network~\cite{Gong18} resulting in white background on try-on images. We use the DensePose coordinates to find the face part, and we additionally used the same segmentation network~\cite{Gong18} to detect hair. As the training progressed, the network has quickly learned to copy the revealed parts through skip-connections, achieving the desired effect. We show examples of garment transfer in \fig{garment}.
We conducted a user-study using 73 try-on samples provided by the authors of~\cite{Raj18}. Participants were given quadruplets of images -- cloth image, person image, our try-on result and result of~\cite{Raj18} and asked to chose which of the try-on images seem more realistic. Since work of~\cite{Raj18} produce only 128$\times$128 images, our results were downsampled. Each sample was assessed by 50 people totalling in 3650 cases, of which our method were preferred in \textbf{57.1\%}.

\subsection{Pose-guided face resynthesis}
\label{subsection:face}

To demonstrate the generality of our idea on texture inpainting, we also apply it to the additional task of face resynthesis. Here, reusing the pipeline used for full body resynthesis, we provide a pair of face images in different poses as a source and a new, unseen view. To estimate the mappings $M_S$ and $M_N$ we use PRNet~\cite{Feng18} --- a state-of-the-art 3D face reconstruction algorithm which provides a full 3D mesh with a fixed number of vertices (43867 in a publicly available version) and triangles (86906). A fixed precomputed mapping from the vertices numbers to their $(u, v)$ texture coordinates is also provided with PRNet implementation. By processing source and target images with PRNet, we obtain estimated $(x, y, z)$ coordinates of a 3D face mesh which leans on an image, such that $(x, y)$ axes are aligned with image axes. We set $(u, v, 1)$ texture coordinates of each vertex as its $(R, G, B)$ color and render a mesh onto an image via Z-buffer, which leaves pixels only visible on a camera view (those not occluded by different faces of a mesh). Similarly to the full body scenario, the obtained rendering for the source view reflects $M_S[x, y]$ mapping, and rendering for the new view reflects $M_N[x, y]$. The pipeline consists of two networks $f$ and $g$ which follow the same architectures as used for the full body view resynthesis. Provided with a source view image and a new view image, the system transfers facial texture from source image onto a pose of a new view image.

\begin{table*}[ht!]
\centering
\small
\begin{tabular}{lllllllllll}
& \vline  & Full body      &            &          &    & \vline  & Face   &    &   &           \\ \hline
& \vline & SSIM$\uparrow$       & MS-SSIM$\uparrow$          & IS$\uparrow$ & LPIPS$\downarrow$ & \vline & SSIM$\uparrow$       & MS-SSIM$\uparrow$          & IS$\uparrow$ & LPIPS$\downarrow$             \\ \hline
\textit{Ours-Full}                             & \vline  & 0.791      &     0.810       &      \textbf{4.46}    &  \textbf{0.169}  & \vline  &  \textbf{0.613}  & \textbf{0.764}    & \textbf{1.834}  & \textbf{0.203}           \\ 
\textit{Ours-NoDeform}                         & \vline &   \textbf{0.797}       &     0.815       &      3.23  & 0.198 & \vline  & 0.609   & 0.758   & 1.819  & \textbf{0.203}  \\  \hline
\textit{RGB inpainting}                        & \vline &   \textbf{0.797}   &     \textbf{0.818}   &      3.02  & 0.198 & \vline  & 0.595   & 0.745   & 1.821  & 0.221    \\ 
\textit{No textures}                           & \vline &   0.796   &     0.812  &    3.295  & 0.202  &  &  &  &  &   \\ \hfill

\end{tabular}
\caption{Ablation study for both \textbf{full body} and \textbf{face} resynthesis. For all algorithms, evaluation is performed based on the same set of validation images. Arrows $\uparrow$, $\downarrow$ tell which value is better for the score --- larger or smaller, respectively. \vspace{-3mm}}
\label{table:ablation}
\end{table*}

\begin{figure*}[ht!]
    \setlength{\tabcolsep}{.1em}
    \centering
    \includegraphics[width=.9\textwidth]{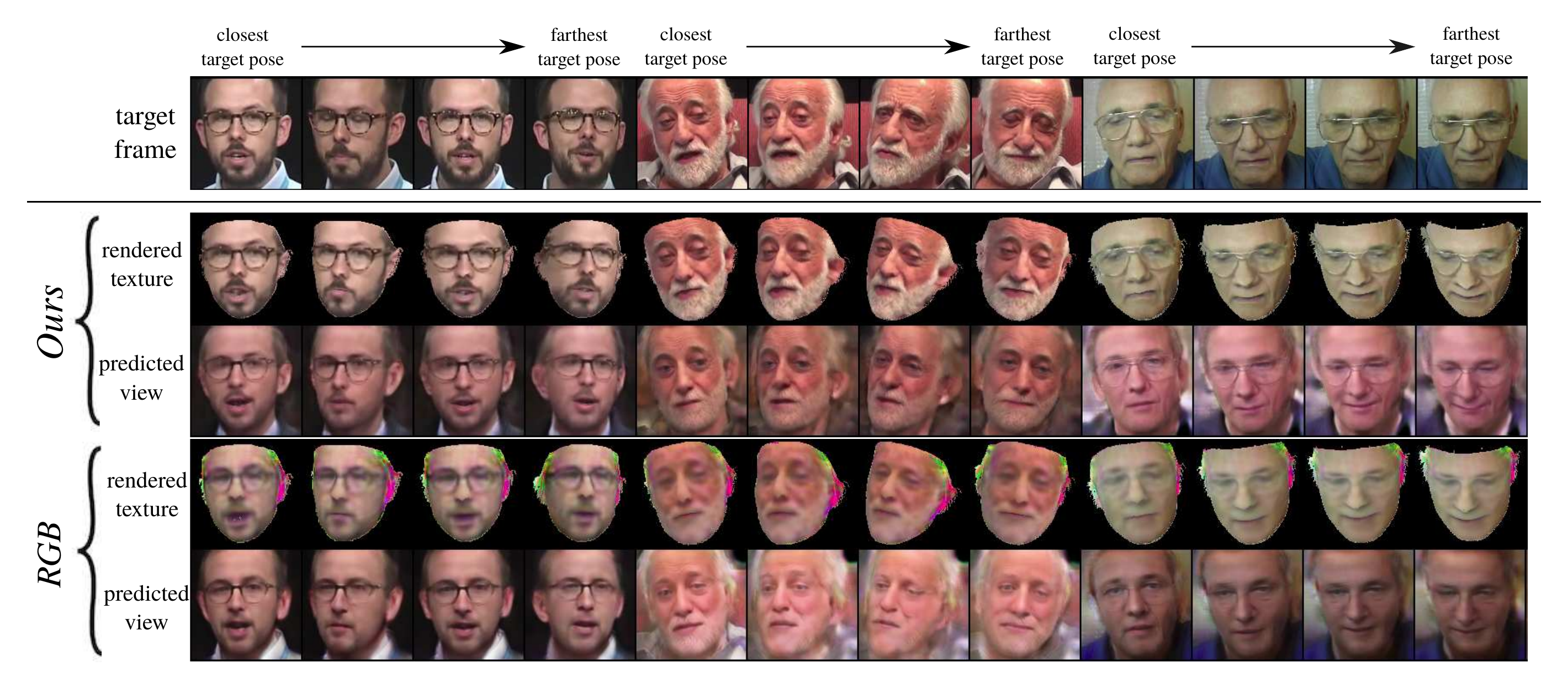}
    \caption{Predictions for several test samples. For each method, we take 3 random subjects, first video frame as a source frame and sample 4 target views according to the 4-quantiles of the pose difference distribution (see testing protocol in Subsection~\ref{subsection:face}). For each subject, source frame is identical to the leftmost target frame shown. In the figure, \textit{rendered texture} refers to the result of warping an inpainted texture onto a new view coordinates, and \textit{predicted view} is a final algorithm output containing the result of texture transfer. Note the differences in sharpness between textures in \textit{Ours} and in \textit{RGB inpainting} and visual quality of their predicted views. \textit{Electronic zoom-in recommended.}}
    \label{fig:faces_ablation}
\end{figure*}


For this subtask, we use 300-VW~\cite{Shen15} dataset of continuous interview-style videos of 114 people taken in-the-wild as a source of training data. Duration of each video is typically around 1 minute and the spatial resolution varies from 480 x 360 to 1280 x 720. Despite that original videos were taken in 25-30 fps, we took each sixth frame of a video in order to speed up the data preparation. Images are preliminarily cropped by a bounding box of 3D face found by PRNet with a margin of 10 pixels and bilinearly resized to a resolution of 128 x 128. Dataset was split into train and validation in proportion of 91 and 23 subjects respectively.

\paragraph{New view resynthesis.} Table~\ref{table:ablation} contains the results of the ablation study, in which we compare three investigated versions of the method (see Subsection~\ref{subsection:pose-guided_image_generation}). The reported values were computed for a subset of 1356 hold out images, collected by a following procedure. For each of 23 videos in the validation set, each $120^{th}$ frame of a video was selected as a source frame. Then, pose orientations of 3D models provided by PRNet were collected for all frames of the video, and angles between pose vector of a source frame 3D model and 3D models of all other frames were calculated. 4 target frames were selected for each source frame as the closest to all of the 4-quantiles of the angles cosine distribution. This way, we test the ability of a model to generalize on target poses both near and far from a source pose (Fig.~\ref{fig:faces_ablation}).

\section{Conclusion}

We have present a new deep learning approach to pose-guided image synthesis. The approach works by estimating the texture of the human body, while a new method for coordinate-based texture inpainting allows to reconstruct detail-rich textures. The reconstructed textures are then used by final resynthesis. The user study suggests that the approach performs well and outperforms state-of-the-art methods  \cite{Siarohin18,Neverova18,Esser18}. We note that for smaller variation of pose, the mapping and estimation of the full texture may be unnecessary, and therefore more direct warping approaches such as \cite{Siarohin18} may be more appropriate under limited changes.



\FloatBarrier
\ifnum\value{page}>8 \errmessage{Number of pages exceeded!!!!}\fi

{\small
\bibliographystyle{ieee}
\bibliography{refs}
}
\end{document}